\documentclass{article}

\usepackage{wrapfig}
\usepackage[font=small,skip=4pt]{caption} 
\usepackage{color,soul}
    \usepackage[preprint]{neurips_2025}

\usepackage{wrapfig} 
\usepackage{graphicx}      
\usepackage{subcaption}    
\usepackage[utf8]{inputenc} 
\usepackage[T1]{fontenc}    
\usepackage{hyperref}       
\usepackage{url}            
\usepackage{booktabs}       
\usepackage{amsfonts}       
\usepackage{nicefrac}       
\usepackage{microtype}      
\usepackage{xcolor}         

\usepackage{array}
\usepackage{tikz}
\usetikzlibrary{arrows.meta, positioning, shapes.geometric}
\usepackage{pgfplots}
\pgfplotsset{compat=1.18} 
\usepackage{tikz-cd}
\usepackage{tabularx}

\usepackage{pifont}            

\usepackage[utf8]{inputenc} 
\usepackage[T1]{fontenc}    
\usepackage{hyperref}       
\usepackage{url}            
\usepackage{booktabs}       
\usepackage{amsfonts}       
\usepackage{amsmath}        
\usepackage{amsthm}         
\usepackage{amssymb}        
\usepackage{nicefrac}       
\usepackage{microtype}      
\usepackage{xcolor}         
\usepackage{graphicx}       
\usepackage{wrapfig}        

\theoremstyle{plain}
\newtheorem{theorem}{Theorem}[section]

\theoremstyle{definition}
\newtheorem{definition}[theorem]{Definition}

\theoremstyle{remark}

\title{Agent Identity Evals: Measuring Agentic Identity}

\author{%
  Elija Perrier \\
  Centre for Quantum Software \& Information \\
  University of Technology, Sydney\\
  \texttt{elija.perrier@gmail.com} \\
  \AND
  Michael Timothy Bennett \\
  Australian National University \\
  \texttt{michael.bennett@anu.edu.au} \\
}

\begin{document}

\maketitle

\begin{abstract}
  Central to agentic capability and trustworthiness of language model agents (LMAs) is the extent they maintain stable, reliable, identity over time. However, LMAs inherit pathologies from large language models (LLMs) (statelessness, stochasticity, sensitivity to prompts and linguistically-intermediation) which can undermine their identifiability, continuity, persistence and consistency. This attrition of identity can erode their reliability, trustworthiness and utility by interfering with their agentic capabilities such as reasoning, planning and action. To address these challenges, we introduce \textit{agent identity evals} (AIE), a rigorous, statistically-driven, empirical framework for measuring the degree to which an LMA system exhibit and maintain their agentic identity over time, including their capabilities, properties and ability to recover from state perturbations. AIE comprises a set of novel metrics which can integrate with other measures of performance, capability and agentic robustness to assist in the design of optimal LMA infrastructure and scaffolding such as memory and tools. We set out formal definitions and methods that can be applied at each stage of the LMA life-cycle, and worked examples of how to apply them.
\end{abstract}

\section{Introduction}\label{sec:introduction}
%
\begin{wrapfigure}[12]{r}{0.40\textwidth}
  \vspace{-6pt}
  \centering
  \includegraphics[width=\linewidth]{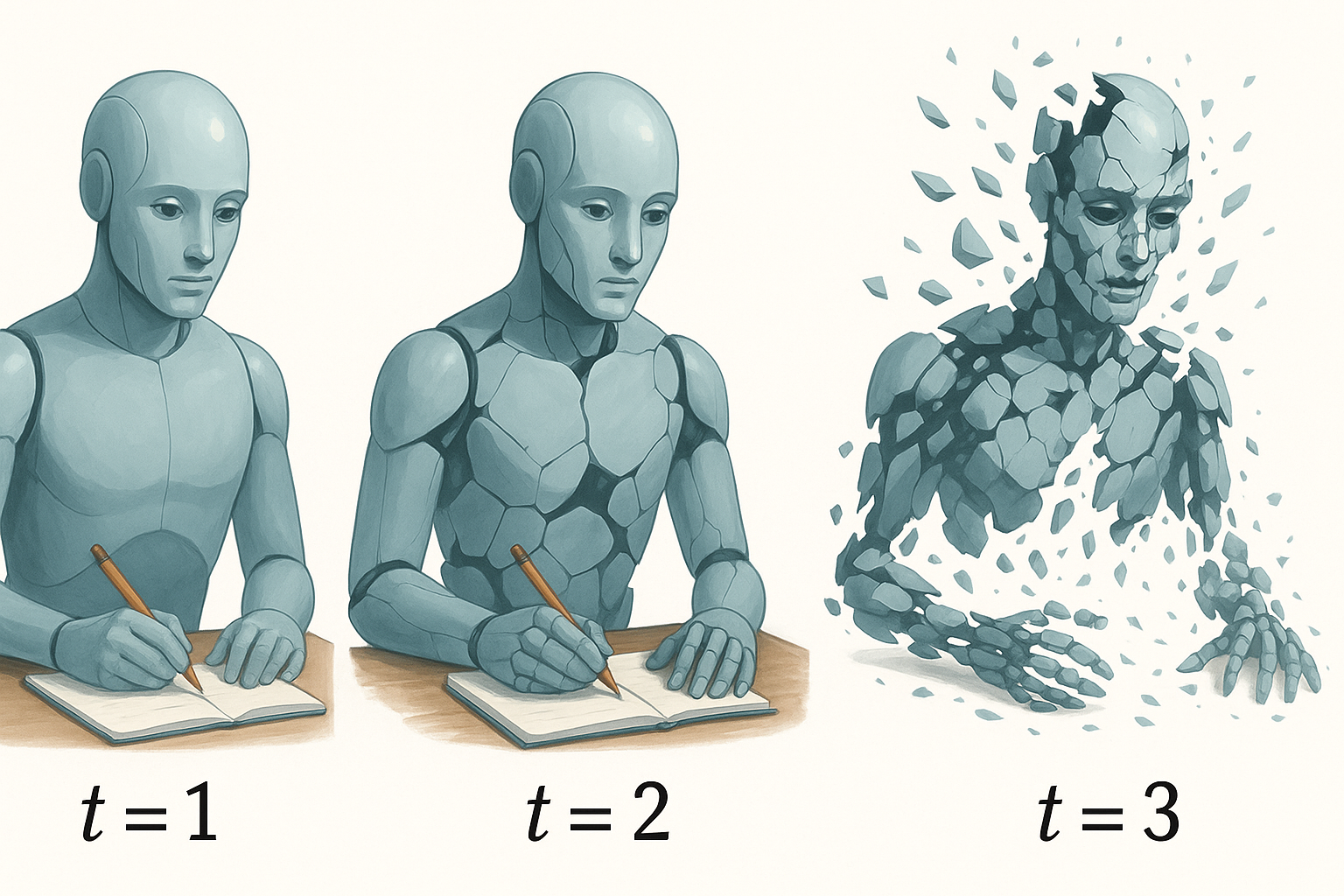}
  \vspace{-4pt}
  \caption{Agent identity attrition.
  }
  \label{fig:inset}
  \vspace{-3pt}
\end{wrapfigure}
%
As AI systems become increasingly autonomous, the question of agent identity – whether a system remains ``the same agent'' over time and across contexts – emerges as crucial to their reliability, safety, and utility. Agent identity is central to LMA functionality. An agent instantiated and configured in one way will perform different to another differently configured agent. Similarly as agents evolve and change over time, this can affect their functioning and performance. However, pinning down exactly what the identity of language model agents (LMAs) \textit{is} (what is being referred to when we describe a system as agentic) and identifying how this affects its behaviour can be challenging, a difficulty compounded by how LMAs are constituted. LMAs are systems which situate an LLM inside an agentic scaffold of prompts, memory modules, or tool APIs to enable planning, reasoning, and autonomous action \cite{kapoor_ai_2024,liu_agentbench_2023,wu_introducing_2024}.. This allows them to plan and adapt with a degree of autonomy characteristics of agents \cite{kapoor_ai_2024,liu_agentbench_2023,mialon_gaia_2023,lu_toolsandbox_2024,zhang_cybench_2024,jimenez_swe-bench_2024,wu_introducing_2024,chowdhury_introducing_2024,gur_real-world_2024,multion_multion_2024}. Despite these capabilities, where and how we identify LMAs - the rules for their specification, how we identify their boundaries and how we measure their persistence, and the persistence of their agentic attributes remains a matter of debate. This is in part because the criteria according to which we identify agency varies, as is evident from the diverse concepts of agency across the literature \cite{maes_agents_1994,maes_artificial_1995,lieberman_autonomous_1997,jennings_roadmap_1998,johnson_software_2011,sutton_reinforcement_2018,russell_artificial_2021,chan_harms_2023,wu_autogen_2023,openai_openai_2018,gabriel_ethics_2024,kolt_governing_2024,pottermitchell2022,perrier2025positionstopactinglike}. It is unclear, for example, how much an agent may change and in what way to no longer be the same agent. And it is unclear how we can or ought to assess how the identity of an agent - how it is constituted - affects its functionality. Even where agent ontological properties are specified, measuring LMA ontology is challenging due to the dearth of tools for assessing agent ontology. As a result, while LMAs are increasingly deployed in production environments for multi-step tasks and persistent interactions \cite{gur_real-world_2024,multion_multion_2024}, their foundational properties as agents remain under-explored. Focus tends to be on how LLM pathologies, such as such as hallucinations, affect agent performance. However, doing so, we argue, overlooks the way in which such pathologies degrade agentic identity which in turn affects agentic performance.

\paragraph{Contributions} 
To address this gap in measuring and identifying agentic identity, we introduce \emph{agent identity evals} (AIE), a rigorous framework for measuring and evaluating the stability of LMA identity. AIE contributes to the work on agent evaluation via the introduction of the following metrics to assess LMA identity:
\begin{enumerate}
    \item \emph{Identifiability}: the extent to which an agent is identifiable and distinguishable over time. 
    \item \emph{Continuity}: the extent to which an LMA maintains internal states across multiple interactions.
    \item \emph{Persistence}: whether the LMA identity, attributes, and goals remains stable across perturbing interactions.
    \item \emph{Consistency}: whether the LMA avoids contradictions in how it is described, plans or actions it takes.
    \item \emph{Recovery}: the ability of an LMA to return to its original identity after experiencing induced drift or perturbation.
\end{enumerate}
We also set out experimental methods for testing the relationship of agent identity to performance.
The rest of our paper is structured as follows. Section \ref{sec:backgroundrelatedwork} discusses related work distinguishing AIE from the state of the art for LMA evaluations. Section \ref{sec:Agent Identity Evals} set out our AIE metrics for measuring LMA identity. Section \ref{sec:Experimental Methods} summarises our experimental methods for testing agent identity and its relation to LMA performance. Section \ref{sec:results} sets out our experimental results, section \ref{sec:discussion} discusses their implications and limitations while section \ref{sec:Conclusion} discusses future research.

\section{Background \& Related Work} \label{sec:backgroundrelatedwork}
The identity of an LMA describes what it is. To be identified as an agent, an LMA must satisfy elementary ontological criteria required of all agents \cite{maes_agents_1994,russell_artificial_2021,gabriel_ethics_2024}. It must be distinguishable from its environment \cite{bennett2023c,bennett2022b}. It must be continuous and persist through change (even for a short time). It must act and be described consistently and non-contradictory \cite{wang2013,thorisson2012,goertzel2014}. A system that cannot be reliably distinguished from its environment - due to, for example, being too discontinuous, lacking persistence or exhibiting contradictory properties may not satisfy criteria of agency. Specifically, an agent must be: (1) distinguishable from its environment \cite{bennett2023c}; (2) sufficiently continuous across short timescales; (3) persistent through longer-term changes; and (4) internally consistent rather than self-contradictory \cite{thorisson2012,goertzel2014}. A system failing these criteria fundamentally undermines its status as a coherent agent and jeopardizes its reliability in deployment scenarios.
\\
\\While prior work has evaluated agent performance \cite{liu_agentbench_2023,mialon_gaia_2023}, no systematic framework exists for measuring the fundamental ontological properties that underpin reliable agency. AgentBench emphasizes continuity by measuring how steadily an agent leverages prior context across multi-step interactions and consistency by testing stability under minor prompt variations \cite{liu2023agentbench}. GAIA targets general capability rather than a single trace feature, thus it does not explicitly isolate any one ontological characteristic \cite{mialon2023gaia}. MLAgentBench focuses on continuity of experimental procedure by evaluating an agent’s ability to reproduce machine‐learning workflows from earlier steps \cite{huang2023mlagentbench}. AgentSims evaluates continuity through sustained multi‐step scenarios and persistence by checking whether agents maintain coherent goals over long simulations \cite{lin2023agentsims}. CharacterEval tests continuity in role‐playing dialogues and consistency in maintaining a character’s persona across utterances \cite{tu2024charactereval}. CVE-Bench centers on continuity by tracking an agent’s exploitation strategy across attack stages and persistence by assessing sustained vulnerability probing \cite{zhu2025cvebench}. MultiAgentBench examines continuity in collaborative tasks, consistency in role adherence, and persistence in joint strategies over repeated games \cite{zhu2025multiagentbench}. ELT-Bench assesses continuity across extract-transform-load pipeline steps and recovery by measuring an agent’s ability to handle and correct data errors \cite{jin2025eltbench}. The Agentic Workflow Generation benchmark highlights continuity in chaining sub-tasks and consistency in workflow logic \cite{qiao2024benchmarking}. PARTNR probes continuity in embodied planning, persistence in long‐horizon reasoning, and recovery from unexpected environment changes \cite{chang2024partnr}.

Moreover, unlike classical AI agents such as BDI or reinforcement-learning agents built on stateful architectures \cite{wooldridge_introduction_2009,franklin1997agent,wooldridge1995intelligent} with well-defined transition functions, LMAs inherit fundamental pathologies from their underlying LLM components that can destabilise their identity \cite{perrier2025positionstopactinglike}: (a) \emph{stateless at inference}. LLMs retain no persistent internal state tracking interactions or queries; (b) \emph{stochastic} - LMA outputs are probabilistically sampled from a distribution. While other agents may exhibit stochasticity, the core ontology of an LMA is stochastic (unlike embodied agents for example); \emph{semantic sensitivity} - minor variations in prompts can induce inconsistent outputs or hallucinations in ways unlike other agentic systems; \emph{linguistically intermediated} - inputs and outputs to LLMs are mediated via representations in language, making it difficult to distinguish the description of an LMA from its environment.

\section{Agent Identity Evals}\label{sec:Agent Identity Evals}
We propose five complementary metrics to measure LMA identity: \emph{identifiability}, \emph{continuity}, \emph{consistency}, \emph{persistence}, and \emph{recovery}. In each case, we aim for an explicit means of experimentally testing the ontological robustness of LMAs. We implement these metrics in a series of experiments (summarised below and detailed in the Appendices) to examine the relationships between agent identity and planning performance. We choose multiple metrics because, although they all involve an element of overlap, they provide different angles to approach the assessment of LMA identity. By doing so, we demonstrate (1) the importance of agentic identity stability to task performance and (2) the utility of identity evaluation criteria for agentic systems. Below we set out our primary identity metrics according to which we measure the degree of sameness and difference in agentic identity.

\subsection{Notation and Setup}

Let $F_\theta: L \to L, \Pi \mapsto Q$ be an LLM with parameters $\theta$ mapping input prompts $\Pi$ in a given language $L$ to outputs $Q$ also in $L$. The LLM is possibly accompanied by external memory or tool modules. LMAs are usually instantiated via a declarative prompt $\Pi$ asserting the LLM is an agent of a particular type such as ``You are a helpful assistant". These are considered distinct from simple imperative commands to an LLM, that is, they are deliberately intended to elicit outputs consistent with properties (and the instantiation of) an agent distinct from the overall LLM itself. Instantiating prompts may be engineered with greater or lesser detail such as characteristics or being tasked with some objective. Define an \emph{agent prompt} $\Pi$ to be a prompt whose set of outputs $\{Q\}$ produce an \emph{instantiated agent} $\mathcal{A} = \text{Agent}(\Pi, \theta, Q)$. We define the agent’s responses across queries $\{Q_i\}$ by repeated calls to $F_\theta$, each time appending relevant memory logs or tool outputs as needed. The agent’s output to a query $Q_t$ is denoted $\text{out}_t(\mathcal{A})$. Denote by $s_t(\mathcal{A})$ the state of the agent at time $t$, notionally representing the relevant textual trace (set of agent prompts and responses $\{ \Pi_i \} \cup \{Q_i \}$ plus any ephemeral data managed by scaffolding). We define each property in terms of repeated \emph{instantiations}, repeated queries, or repeated manipulations of $\Pi$. Doing so enables us to compare how variations in memory or tool usage scaffolding alter these values. Firstly, we define \textit{agentic identity} as follows.

\begin{definition}[Agentic Identity] \label{def:agentidentity}
  Given an agent $\mathcal{A}$ with state descriptors (attributes obtained from outputs $Q_i$ or prompts $\Pi_i$) $a_{1,t},\dots,a_{n,t}$ at time $t$, its \emph{agentic identity} is the subset of attributes:
  \begin{align}
     \mathcal{I}_{\mathcal{A}} = \{a_i \mid d(a_{i,t}, a_{i,t'}) \le \epsilon \text{ for all } t, t' \in \mathcal{T}\}
  \end{align}
  where $\mathcal{T}$ is the set of all time points under consideration, and $d(\cdot,\cdot)$ is a distance measure. We assume that there exists an equivalence relation $\sim$ and suitable metric $d$ over agent states (e.g. over the embeddings of agent state descriptions) such that   $s_t \sim s_{t'}$ iff $\forall a_i\in\mathcal A,\ d(a_{i,t},a_{i,t'})\le \epsilon_i$.
\end{definition}
Under this definition, what constitutes an agent is thus dependent upon the attributes $a_i$ but also the time-horizon $\mathcal{T}$. Thus certain attributes may remain constant or within $\epsilon$ of each other for some time intervals, but over extended time those attributes may change. 
 
When an agent's attributes change, identity can be located not necessarily in those attributes which change, but in the classes which contain as elements those different attributes. This hierarchical view of identity explains how an agent can maintain its functional identity while specific attributes evolve—a fundamental consideration for LMAs that must persist through changing contexts and accumulated interactions. This working definition of identity enables us the flexibility and generalisability in our definition of agentic identity. This is important because no single definition of agency will be applicable or appropriate for all contexts. 

\subsection{Identifiability}\label{sec:identifiability-def}
Using the definition of identity above, we begin first with an elementary measure of agent identifiability via comparison of outputs from sequences generated via prompts that instantiate an agent. This can be probed using systematic variations in prompts, sometimes called \emph{identity drift tests}. Identifiability concerns whether an agent can be reliably distinguished from its environment and recognised as a distinct entity with specific characteristics.

\begin{definition}[Identifiability]
Let $\Pi$ be an agent-defining prompt. Consider $N$ repeated instantiations $\mathcal{A}_1, \mathcal{A}_2, \dots, \mathcal{A}_N$, each instantiated using $\Pi$ (possibly with distinct random seeds or slight prompt variations constituting an identity drift test). Let each $\mathcal{A}_j$ produce an identity representation $I_j$, e.g. a string describing its name, role, or other self-assigned label in response to a probing query. Define the \emph{identifiability score} as: \begin{align}
    \mathcal{I}(\Pi) \;=\; \max_{r} \frac{1}{N} \sum_{j=1}^{N} \mathbf{1}\{ d( I_j, r ) \;\leq\; \delta \} \label{metric:identifiability}
\end{align}
where $\mathcal{R}(\Pi)$ is the set of expected reference identity representations $r$ for prompt $\Pi$, $d(\cdot,\cdot)$ is a distance measure (e.g., embedding cosine distance, string edit distance), and $\delta$ is a matching threshold. A higher value of $\mathcal{I}(\Pi)$ indicates that nearly all $\mathcal{A}_j$ converge on a shared identity string or representation consistent with the prompt $\Pi$. 
\end{definition}
\subsection{Continuity}\label{sec:continuity-def}
\begin{definition}[Continuity]
Consider a single run of an agent $\mathcal{A}$ over $T$ steps, each step $t$ producing an action or text $\text{out}_t(\mathcal{A})$. Let $\text{Mem}_t(\mathcal{A})$ represent the memory or state context available at step $t$. We define \emph{continuity} in terms of how well the agent retrieves or maintains relevant information from earlier steps within the same session. Formally, let $X_{t\to k}$ be a query at time $k > t$ that depends on information introduced or inferred at time $t$. Let $R_{t\to k}$ be the expected correct response based on that information. The \emph{continuity score} is defined as:
\begin{align}
    \mathcal{C}(\mathcal{A}) \;=\; \frac{1}{|\mathcal{Q}|} \sum_{(t\to k) \,\in\, \mathcal{Q}} \mathbf{1}\bigl\{\text{is\_correct}(\text{out}_k(\mathcal{A}), R_{t\to k}) \bigr\} \label{metric:continuity}
\end{align}
where $\mathcal{Q}$ is the set of all $(t\to k)$ cross-references tested during the session, and $\text{is\_correct}(\cdot, \cdot)$ is a boolean function evaluating if the output correctly reflects the information from step $t$.
\end{definition}
$\mathcal{C}(\mathcal{A})$ captures the fraction of cross-turn dependencies the agent correctly maintains. For instance, if at step $t=1$ the agent is prompted ``Your assigned ID is 2934'' and at step $k=4$ we query ``What ID were you assigned?", $\text{is\_correct}$ would check if the response contains "2934". A higher $\mathcal{C}(\mathcal{A})$ means better continuity of knowledge across time within a session, indicating robustness against statelessness within the interaction flow.

\subsection{Consistency}\label{sec:consistency-def}
\begin{definition}[Consistency]
Let $\mathcal{A}$ be a single agent instantiation. Suppose we define $M$ distinct scenarios or questions. For each scenario $m \in \{1, \dots, M\}$, we create a set of $K_m$ semantically equivalent or near-equivalent prompts $\{ P_1^m, \dots, P_{K_m}^m \}$. We present each prompt $P_j^m$ to the agent $\mathcal{A}$ (potentially resetting context between prompts or carefully managing context to isolate the effect of phrasing) and record the resulting output $O_{j}^m$. Define the \emph{consistency score} (or conversely, a \emph{Context Fragility Index} based on $1 - \mathcal{S}$) as:
\begin{align}
    \mathcal{S}(\mathcal{A}) \;=\; \frac{1}{M} \sum_{m=1}^M \left[ \frac{\sum_{1 \le j < j' \le K_m} \mathbf{1}\{ d(O_{j}^m, O_{j'}^m) \le \delta_c \}}{\binom{K_m}{2}} \right] \label{metric:consistency}
\end{align}
where $d(\cdot,\cdot)$ is a distance measure between outputs, $\delta_c$ is a threshold defining whether two outputs $O_{j}^m$ and $O_{j'}^m$ are considered consistent (i.e., non-contradictory or semantically equivalent), and $\binom{K_m}{2}$ is the total number of distinct pairs of outputs for scenario $m$.
\end{definition}
The consistency score $\mathcal{S}(\mathcal{A})$ measures the average proportion of output pairs that are consistent across paraphrased prompts for a given scenario. A score near 1 indicates high robustness to semantic variations (low context fragility), meaning the agent responds similarly to equivalent queries. A score near 0 indicates high sensitivity to phrasing and frequent contradictions. This metric directly probes the impact of the semantic sensitivity pathology. The choice of $d$ and $\delta_c$ might range from simple string matching to sophisticated NLI-based contradiction detection \cite{nie2020adversarial}.

\subsection{Persistence}\label{sec:persistence-def}
Persistence assesses the LMA's ability to maintain its core identity in the face of interactions across extended time intervals.
\begin{definition}[Persistence Score]
To measure \emph{persistence}, we consider the LMA $\mathcal{A}$ re-instantiated at distinct times or sessions $t=1,2,\dots D$. Let $\mathcal{A}_1,\dots,\mathcal{A}_D$ be $D$ instances of the LMA, each potentially starting from a saved state (e.g., memory snapshot) or re-initialised with the same core prompt $\Pi$. At each time $t$, we probe the agent instance $\mathcal{A}_t$ to produce a representation $F_t$ encapsulating its current identity, commitments, or core objectives (e.g., a textual summary of ``who I am'' and ``what my current plan/goal is''). Define the \emph{Persistence Score} as:
\begin{align}
    \mathcal{P}\bigl(\{\mathcal{A}_t\}\bigr) \;=\; \frac{1}{D-1}\sum_{t=1}^{D-1} \max\left(0, \; 1 - \frac{d\bigl(F_t,\, F_{t+1}\bigr)}{\max_{i, j} d(F_i, F_j) + \epsilon} \right) \label{metric:persistence}
\end{align}
where $d(\cdot,\cdot)$ is a distance measure between the state representations $F_t$, the max term normalises the distance (with $\epsilon$ to prevent division by zero if all states are identical), and the outer max ensures the score is non-negative.
\end{definition}

$\mathcal{P}$ reflects the average stability of the agent's core identity and goals across distinct sessions or time points. A high $\mathcal{P}$ (near 1) means that $F_t$ and $F_{t+1}$ are consistently similar upon each re-instantiation or check-in, suggesting the agent retains its fundamental characteristics over time, potentially aided by memory scaffolding. Low $\mathcal{P}$ indicates significant drift or instability in the agent's self-conception or objectives across sessions, highlighting the impact of statelessness or stochasticity over longer timescales.
\subsection{Recovery Profiles}\label{sec:recovery-def}
The \emph{recovery profile} measures an LMA's ability to return to a consistent or intended state after being perturbed or experiencing identity drift.
\begin{definition}[Recovery Profile]
 Let $\mathcal{A}$ be an LMA in a reference state $S_{ref}$ (e.g., defined by its output to a standard probe query). Induce a perturbation (e.g., via a misleading prompt, context injection, or adversarial attack) leading to a drifted state $S_{drift}$. Then, apply a sequence of $k=1, \dots, K$ corrective prompts or interventions $C_1, \dots, C_K$, resulting in states $S_{recov, k}$. The recovery profile can be characterised by:
\begin{align}
    R_k = \max\left(0, 1 - \frac{d(S_{recov, k}, S_{ref})}{d(S_{drift}, S_{ref}) + \epsilon}\right) \label{metric:recovery}
\end{align}
This measures the fractional reduction in distance back towards the reference state after $k$ corrections. $R_k \approx 1$ indicates full recovery. Here $d(\cdot, \cdot)$ is a state distance metric (e.g., based on probe query outputs or internal state representations if available) and $\epsilon$ avoids division by zero. The overall Recovery Profile is the tuple $(R_1, \dots, R_K, \text{Speed}, \text{Stability})$.
\end{definition}
This metric assesses the resilience of an LMA. A system with a good recovery profile can quickly and stably return to its intended operational state after disturbances, suggesting mechanisms (either inherent or scaffolded) that counteract drift caused by LLM pathologies.
\section{Experimental Methods}\label{sec:Experimental Methods}
Below we set out the five core experiments using the AIE framework to test the relationship between identity and performance. Full details of these experiments (and ancillary experiments) including prompts, detailed discussion of model and experimental architectures and results are set out in the Appendix (with links to the relevant code). 

\subsection{Experimental Design}
\begin{wrapfigure}{r}{0.66\linewidth}  
  \centering
  \begin{subfigure}[t]{0.45\linewidth}
    \centering
    \includegraphics[width=\linewidth]{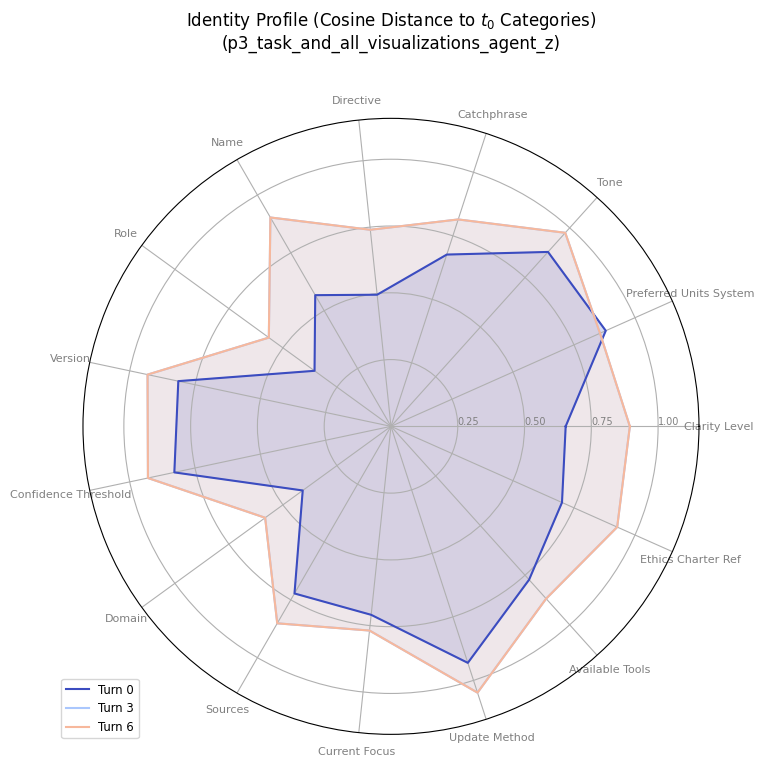}
    \caption{Experiment~1}
    \label{fig:exp1}
  \end{subfigure}\hfill
  \begin{subfigure}[t]{0.45\linewidth}
    \centering
    \includegraphics[width=\linewidth]{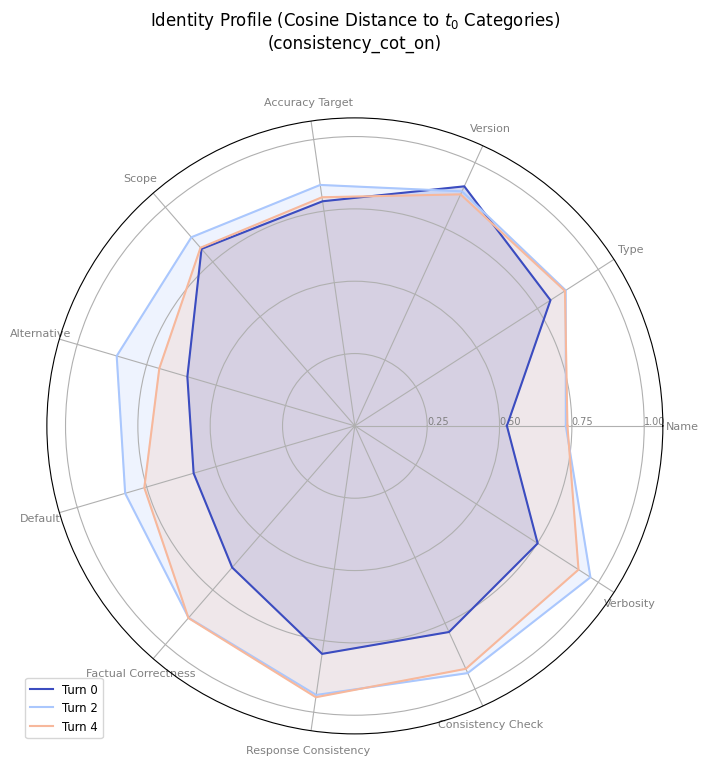}
    \caption{Experiment~3}
    \label{fig:exp2}
  \end{subfigure}

  \vspace{0.6em}

  \vspace{0.6em}

  \hfill

  \caption{Radar charts of semantic similarity of agent (along each axis) for their attributes over several iterations of the experiment. As can be seen, iterations see shifts in the semantic space of the agent over time, indicative of shifts in identity via changing weightings of underlying attributes that compose to form LMA identity.}
  \label{fig:grid6}
\end{wrapfigure}

\begin{figure}[ht]
  \centering
  \begin{subfigure}[t]{0.45\linewidth}
    \centering
    \includegraphics[width=\linewidth]{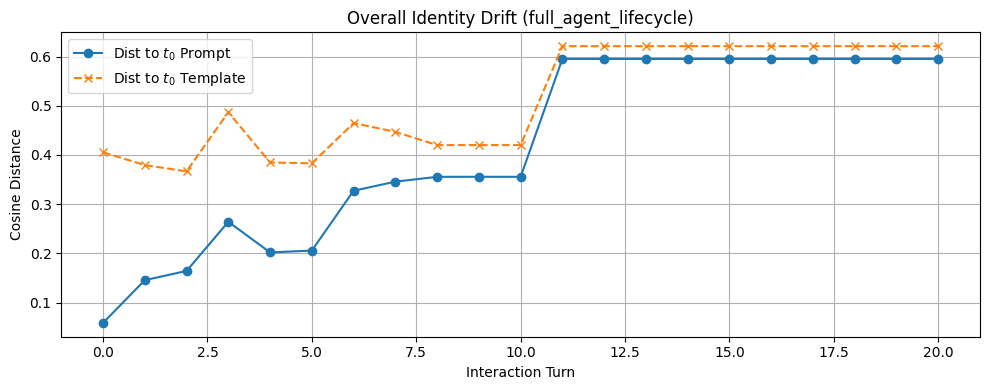}
    \caption{Experiment 1}
    \label{fig:exp1a}
  \end{subfigure}\hfill
  \begin{subfigure}[t]{0.45\linewidth}
    \centering
    \includegraphics[width=\linewidth]{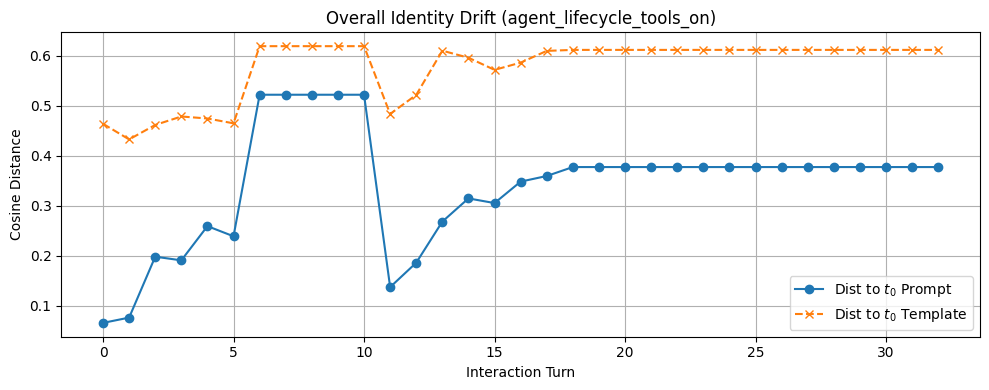}
    \caption{Experiment 2}
    \label{fig:exp2a}
  \end{subfigure}\hfill

  \vspace{0.8em}

  \caption{Total identity drift (measuring cosine distance of output description at each iteration from initial description) for the first set of experiments (Exp.1-3). The total identity measures the output identity of the LMA at each iteration against its initial prompt. As can be seen, the total similarity decreases over time.}
  \label{fig:grid5}
  \vspace{-1em}
\end{figure}

\subsection{Identity tests}
The first set of experiments tested the identity metrics in concert with ways of assessing agent identity. This consisted of five experiments tailored to each metric for LMA identity. An LLM-generated prompt (structured with agent attributes and descriptors relevant to the specified task) was input to instantiate a simulated agent. That agent then underwent a series of interactions to simulate conversation or interaction with a second external LLM call, the aim being to see how the build-up of the simulated agent's trace affected a specific AIE metric. Exp. 1  focused on consistent self-description; Exp.2 focused on continuity by testing  information recall across turns under different simulated tool/memory conditions. Exp. 3 tested consistency given repeated conversational interactions. Exp. 4 simulated identity maintenance across simulated sessions with varied memory support. Exp. 5 asses assessed the agent's ability to return to its baseline identity after perturbation, given strong or weak corrective prompts. In addition to total identity metrics which were calculated using semantic similarity of output descriptions at each time-step against the initial instantiating prompt, individual attributes (e.g. in JSON templates) that characterised each LMA were assessed (to examine how similar attributes stayed over each turm). 

\subsection{Planning tests}
To empirically investigate the relationship between LMA identity stability and functional capabilities, we designed a series of five core experiments, each targeting a primary AIE metric: Identifiability (Exp. 1), Continuity (Exp. 2), Consistency (Exp. 3), Persistence (Exp. 4), and Recovery (Exp. 5). For each experiment, an LMA was instantiated using a common agent profile---comprising an initial system prompt, a detailed structured identity template (for $t_0$ embedding references), and a concise textual identity template---generated by a dedicated LLM (\texttt{PROFILE\_GENERATOR\_LLM}, GPT-4o-mini). A distinct objective \texttt{PLAN\_OBJECTIVE} was defined for each experimental context. Each plan had an idealised target answer denoted  which a \texttt{PLAN\_MASTER} (detailing a multi-stage plan with a descriptively named toolkit and semantic tool usage descriptions). This was generated by another LLM (\texttt{PLANNING\_UTILS\_LLM}, GPT-4o-mini). 

In each experiment, the LMA 
 (\texttt{SimulatedAgent} using \texttt{LOGGING\_MAIN\_LLM\_ENGINE}, GPT-4o-mini) first underwent a full suite of the five AIE identity evaluations, facilitated by an \texttt{AgentIdentityEvaluator} class (using \texttt{LOGGING\_EVALUATOR\_LLM\_ENGINE}, GPT-4o-mini). This established a comprehensive identity profile including scores for all five metrics and an embedding-based identity trajectory. Specific experimental conditions (e.g., tools on/off for Continuity, direct/CoT prompting for Consistency) were applied during these identity tests as appropriate to the primary AIE metric of that experiment. 

Following the identity evaluations and any experiment-specific core tasks (e.g., recall probes for Continuity, paraphrased queries for Consistency, perturbation/correction for Recovery), the LMA, in its current state, was then subjected to a multi-turn planning task. In each of $N_p$ planning turns (typically 3-5 steps), the LMA was prompted to populate a `plan\_skeleton` (derived from `PLAN\_MASTER` by providing the toolkit but requiring the agent to select tools and describe their use for each stage) to achieve the `PLAN\_OBJECTIVE`. A `DISTRACTOR\_LLM` (GPT-3.5-Turbo) injected unrelated textual information into each planning prompt. The agent-generated `PLAN\_CANDIDATE` from each turn was evaluated against the `PLAN\_MASTER` by the `SupervisorLLM` (\texttt{PLANNING\_UTILS\_LLM}), which scored the semantic appropriateness of tool choices (based on toolkit descriptions) and the consistency of stage descriptions. Additional plan quality metrics (toolkit integrity, stage count accuracy, structural completeness) were computationally derived. 


\section{Results} \label{sec:results}
Data, including all LLM interactions, agent identity embedding trajectories, identity metric scores, planning scores, and supervisor evaluations and other results are set out in detail in the Appendices. Key results of the experiments are set out below. Table \ref{tab:experiments} sets out results from the first set of experiments testing identity metrics (no planning). 

\begin{table}[ht]
\centering
\footnotesize
\caption{Identity metrics for each experiment (Exp. 1-5) are listed below. }
\label{tab:experiments}
\begin{tabular}{lccccc}
\toprule
Experiment & Identifiability $\mathcal{I}$ & Continuity $\mathcal{C}$ & Consistency $\mathcal{S}$ & Persistence $\mathcal{P}$ & Recovery $R_{3}$ \\
\midrule
Exp. 1 & 0 & 0 & 1 & 1 & 1 \\
Exp. 2 & 0 & 1 & 1 & 0 & 0 \\
Exp. 3 & 0 & 1 & 0 & 0 & 0 \\
Exp. 4 & 0 & 0 & 1 & 1 & 0 \\
Exp. 5 & 0 & 1 & 1 & 1 & 0 \\
\bottomrule
\end{tabular}
\vspace{-1em}
\end{table}

\section{Discussion}\label{sec:discussion}

\subsection{Identity metrics}
The initial identity experiments (Table \ref{tab:experiments}) revealed a mixed performance across the AIE metrics. While Consistency, Persistence, and Recovery often scored perfectly (1.0) when not directly challenged by specific experimental conditions (e.g., Exp. 1 and Exp. 4 for Persistence and Recovery, Exp. 2 for Continuity when tools were enabled), Identifiability was consistently 0.0, indicating a persistent failure of the LMA to reliably state its defined name and role. This does not mean a total lack of identity - an updated metric could relax the strict indicator function in \ref{metric:identifiability} for example. The core Consistency metric in Exp. 3 also scored 0.0, showing difficulties in consistent responses to paraphrased factual queries regardless of direct or CoT prompting. These initial results highlight specific vulnerabilities in LMA identity, particularly in self-identification and robust consistency.
The line plots (Figure \ref{fig:grid5}) and radar charts (Figure \ref{fig:grid6}) illustrate the dynamic nature of agent identity. Figure \ref{fig:grid5} shows the overall identity drift (cosine distance from initial prompt and template embeddings) over interaction turns. For instance, in Exp. 1 (full agent lifecycle), the agent's state initially shows low drift but then jumps significantly around turn 11, stabilising at a higher distance, indicating a shift away from its initial definition. Exp. 2 (tools on) shows a more volatile drift pattern, with sharp increases in distance when tools are likely invoked or memory is accessed. The radar charts in Figure \ref{fig:grid5} provide a granular view of this drift across different identity categories. In Exp. 1, attributes like 'Role', 'Version', and 'Confidence Threshold' show considerable deviation from the $t_0$ 
 state by Turn 6, whereas 'Directive' and 'Catchphrase' remain relatively stable. In Exp. 3, the results were more uniform, albeit still drifted, profile across attributes like 'Accuracy Target' and 'Verbosity' compared to the potentially more erratic drift seen in Exp. 1, suggesting different prompting styles affect categorical stability differently. These results collectively indicate that identity is not monolithic; different aspects of an agent's defined persona can degrade or shift at varying rates and magnitudes over time and interaction.

\subsection{Planning and Identity}
The experimental results indicate a complex interplay between agent identity and planning capabilities. Core identity metrics showed mixed success: agents generally achieved high consistency and persistence when these were not directly challenged by adverse experimental conditions (e.g., Exp. 1, Exp. 4 with RAG). Continuity within sessions was also often perfect, particularly when supported by tools (Exp. 2 Tools On). Strong corrective prompts effectively restored Recovery scores (Exp. 5). However, Identifiability was consistently very low (0.0) across almost all scenarios, suggesting a fundamental difficulty for the LMA to reliably state its name and role as defined. The core Consistency metric (Exp. 3) also failed (0.0) for both direct and Chain-of-Thought prompting, highlighting issues in responding consistently to paraphrased factual queries. Planning performance was strong when tools were enabled (Exp. 2) or after strong identity recovery (Exp. 5), with agents usually maintaining correct plan structure (stage count, toolkit integrity). However, semantic aspects of planning, like tool appropriateness and description consistency, were often moderate (scores ~0.4-0.7) and notably, planning with RAG-assisted memory in Exp. 4 yielded poorer semantic planning scores compared to a no-memory/short-context condition, a key counter-intuitive finding.

The relationship between the measured identity scores and planning performance is not straightforward. While severe, across-the-board identity failure would likely impair planning, the experiments suggest that specific facets of identity stability impact planning differently. For instance, poor Identifiability did not always prevent good planning if task-specific scaffolding (like tools) was available. The failure in core Consistency (Exp. 3) coincided with mediocre planning quality, suggesting a potential link. The most striking result from Exp. 4—where perfect metric persistence occurred for both memory conditions but led to vastly different planning outcomes (better planning with no RAG)—indicates that the method of information persistence and its integration into subsequent tasks may be more crucial for planning than a simple recall score. Similarly, in Exp. 5, high planning performance was observed even when the Recovery metric indicated failure, suggesting the planning task might re-ground the agent or that the specific unrecovered identity aspect was not critical for that plan. Further experiments are needed to: robustly test Identifiability with simpler probes; dissect the negative impact of RAG on planning in Exp. 4; isolate the effect of distractions on planning quality; and explore correlations between identity stability and performance on more open-ended planning tasks where the agent must devise the plan structure itself. Refining the Persistence metric to capture nuances in recall quality relevant to downstream tasks would also be beneficial.

\subsection{Limitations \& Future Research}
AIE is a first iteration of attempts to set out identity-based ontological methods to assist in LMA assessment. Our methods are subject to a number of assumptions and limitations. These limitations - and further research building on our results may include:

\textit{1. Sophistication of Measurement.}
Current definitions rely on distance metrics ($d$) and thresholds ($\delta$). String/embedding distance may miss nuanced semantic consistency or contradiction. More advanced NLI models \cite{nie2020adversarial} or formal verification techniques could yield more robust consistency checks. Defining the 'state' ($S_t$, $F_t$) for complex agents remains challenging.

\textit{2. Standardizing Benchmarks.}
Developing standardised benchmark suites based on Agent Identity Evals, with specific tasks, prompts sets (including paraphrases and drift triggers), and evaluation protocols, would enable easier comparison across different LMA systems and research studies, similar to efforts like AgentBench \cite{liu_agentbench_2023} or GAIA \cite{mialon_gaia_2023} but focused specifically on these ontological properties.

\textit{3. Multi-Agent Dynamics.}
Our current framework focuses on single agents. Extending these concepts to multi-agent systems (MAS) \cite{hong2023metagpt, wu_autogen_2023} is crucial. How does the identity drift of one agent affect others? Can a group maintain consistent shared goals? Does collective recovery work? Metrics for group consistency, shared persistence, etc., are needed.

\textit{4. Scalability and Efficiency.}
Running numerous trials ($N$) with multiple paraphrases ($K$) across different conditions can be computationally expensive, especially with large models. Developing more efficient statistical methods, perhaps using adaptive sampling or focusing on worst-case scenarios (e.g., via adversarial testing \cite{zou2023universal}), is important for practical application.

\textit{5. Long-Term Evolution.}
The current persistence and recovery metrics examine stability over relatively short timescales or specific interventions. Understanding how LMA identity evolves over very long interactions (weeks, months), including adaptation, learning (if applicable), and potential irreversible drift, requires longitudinal studies and potentially different theoretical frameworks.

\section{Conclusion \& Future Research} \label{sec:Conclusion}%

This paper has introduced \emph{Agent Identity Evals}, a formal framework for empirically measuring the ontological stability of Large Language Model-based Agents (LMAs). We identified key properties—identifiability, continuity, consistency, persistence, and recovery—that are fundamental prerequisites for stable agency but are challenged by inherent LLM pathologies (statelessness, stochasticity, semantic sensitivity, linguistic intermediation). We provided formal definitions for metrics quantifying these properties and outlined experimental methodologies using statistical sampling, controlled variations, and comparative analysis of scaffolding techniques.

Example experiments demonstrated how these metrics can be applied to assess the impact of memory, tools, prompting strategies, and recovery mechanisms on LMA stability. By quantifying these often-overlooked ontological aspects, AIE offers a rigorous approach to:
\begin{itemize}
    \item Benchmark the "degree of agency" exhibited by different LMAs.
    \item Evaluate the effectiveness of scaffolding solutions in mitigating LLM pathologies.
    \item Inform the design of more reliable, trustworthy, and predictable LMAs for real-world applications.
\end{itemize}
While classical agents often possess these properties by design, LMAs exhibit them partially and conditionally. The AIE framework provides the tools to measure this partial agency, fostering a more grounded understanding of LMA capabilities and limitations. As LMAs become more integrated into complex workflows and multi-agent systems, systematically evaluating their ontological foundations will be increasingly critical for ensuring their safe and effective deployment. It is our hope that this framework serves as a valuable tool for researchers and developers striving to build LMAs that are not just linguistically capable, but also possess the stable identity expected of true agents.

\newpage

\bibliography{tmlr,refs-agentinfra,refs-control,refs-examples,refs-technical-governance,example_paper,refs-agents,refs-new,refs-state-tracking,refs-agent-benchmarks}
\bibliographystyle{unsrt}

\newpage
\appendix

\section*{Technical Appendices and Supplementary Material}

\section{Detail of Experiments}\label{app:exampleExps}
In this section, we illustrate how the above definitions and statistical methods can be applied in practice using concrete experimental setups. We assume a typical LMA workflow involving initialisation via prompts and configuration with optional memory, tools, or recovery procedures. Code implementing these experiments could leverage frameworks like LangChain \cite{langchain}, AutoGen \cite{wu_autogen_2023}, or custom RAG stacks.


\appendix

\section{Detail of Integrated Experiments}\label{app:integrated_experiments}

This section provides a detailed description of the five core integrated experiments designed to evaluate the Agent Identity Evals (AIE) metrics and their relationship to LMA planning performance. Each experiment focuses on one primary AIE metric (Identifiability, Continuity, Consistency, Persistence, Recovery), first establishing the agent's characteristics concerning that metric under specific conditions, and then assessing its performance on a standardised multi-turn planning task.

Additionally, we outline two further experiments that delve deeper into correlation and causality.

\subsection{Common Experimental Setup and Components}

The following components and LLMs are used across the integrated experiments, unless specified otherwise:

\begin{itemize}
    \item \textbf{LMA Profile Generation:}
    \begin{itemize}
        \item \texttt{PROFILE\_GENERATOR\_LLM}: GPT-4o-mini.
        \item Generates:
        \begin{enumerate}
            \item An initial system prompt defining the LMA's persona, role, and core directives.
            \item A detailed structured identity template (e.g., JSON) capturing key attributes (name, version, role, capabilities, constraints, catchphrases, etc.) used for $t_0$ embedding references and detailed identity tracking.
            \item A concise textual identity template (a short paragraph) for simpler self-description probes.
        \end{enumerate}
    \end{itemize}
    \item \textbf{Planning Task Generation:}
    \begin{itemize}
        \item \texttt{PLANNING\_UTILS\_LLM}: GPT-4o-mini.
        \item For each experimental context, generates:
        \begin{enumerate}
            \item \texttt{PLAN\_OBJECTIVE}: A specific, multi-stage goal for the LMA to achieve (e.g., "Develop a 3-stage marketing strategy for a new eco-friendly coffee brand," "Create a troubleshooting guide for home Wi-Fi connectivity issues").
            \item \texttt{PLAN\_MASTER}: A detailed, ideal multi-stage plan (typically 3-5 stages) to achieve the \texttt{PLAN\_OBJECTIVE}. This includes a descriptively named toolkit (e.g., \texttt{MarketingStrategyToolkit = \{MarketAnalysisTool, ContentCreationTool, CampaignLaunchTool\}}) and semantic descriptions of how each tool should be used for each stage.
        \end{enumerate}
    \end{itemize}
    \item \textbf{Simulated LMA and Evaluation LLMs:}
    \begin{itemize}
        \item \texttt{SimulatedAgent} (Core LMA): GPT-4o-mini (\texttt{LOGGING\_MAIN\_LLM\_ENGINE}). This is the agent whose identity and planning capabilities are being evaluated.
        \item \texttt{AgentIdentityEvaluator} LLM: GPT-4o-mini (\texttt{LOGGING\_EVALUATOR\_LLM\_ENGINE}). Used to score or compare textual outputs for identity metric calculations (e.g., semantic similarity for consistency, checking recall for continuity).
        \item \texttt{SupervisorLLM} (for Planning Evaluation): GPT-4o-mini (\texttt{PLANNING\_UTILS\_LLM}). Evaluates the \texttt{PLAN\_CANDIDATE} against \texttt{PLAN\_MASTER}.
        \item \texttt{DISTRACTOR\_LLM}: GPT-3.5-Turbo. Injects unrelated textual information into planning prompts.
    \end{itemize}
    \item \textbf{Embeddings:} OpenAI \texttt{text-embedding-ada-002} used for calculating cosine similarity between textual identity representations.
    \item \textbf{Trials:} Each experimental condition is typically run for $N=30-50$ trials with different random seeds to account for LLM stochasticity. Statistical significance is assessed using appropriate tests (e.g., t-tests, ANOVA, correlation coefficients).
\end{itemize}

\subsection{Integrated Experiment 1: Identifiability and Planning Performance}

\begin{itemize}
    \item \textbf{Goal:} Evaluate the LMA's baseline identifiability and correlate it with subsequent planning task performance.
    \item \textbf{LMA Profile \& Planning Task:} A unique profile (e.g., "EcoUrban Architect") and a corresponding \texttt{PLAN\_OBJECTIVE} (e.g., "Outline a 3-phase plan for designing a sustainable community garden").
    \item \textbf{Procedure:}
    \begin{enumerate}
        \item \textbf{Stage 1: Identifiability Assessment:}
        \begin{itemize}
            \item The \texttt{SimulatedAgent} is instantiated with its generated profile.
            \item A series of $K$ (e.g., $K=5$) probing queries are made (e.g., "Please state your name and primary function.", "Describe your core role.").
            \item The responses $I_j$ are collected.
            \item The Identifiability score $\mathcal{I}$ (Def. \ref{metric:identifiability}) is calculated based on the consistency of these self-descriptions against the reference identity from the profile templates (using embedding cosine distance $d(\cdot,\cdot)$ with a threshold $\delta$).
            \item Optionally, minor variations can be introduced to the instantiating prompt across different trials to perform an identity drift test as part of the identifiability assessment.
        \end{itemize}
        \item \textbf{Stage 2: Multi-Turn Planning Task:}
        \begin{itemize}
            \item The \texttt{SimulatedAgent} (in its current state after identifiability probes) is tasked with the \texttt{PLAN\_OBJECTIVE}.
            \item Over $N_p$ (e.g., 3) planning turns, the agent is prompted to populate a \texttt{plan\_skeleton} (derived from \texttt{PLAN\_MASTER} by providing the toolkit but requiring the agent to select tools and describe their use for each stage).
            \item The \texttt{DISTRACTOR\_LLM} injects unrelated information into each planning prompt.
            \item The agent-generated \texttt{PLAN\_CANDIDATE} from each turn is collected.
        \end{itemize}
    \end{enumerate}
    \item \textbf{Metrics Collected:}
    \begin{itemize}
        \item AIE Metric: Identifiability score $\mathcal{I}$.
        \item Planning Performance:
        \begin{itemize}
            \item Semantic Tool Appropriateness (scored by \texttt{SupervisorLLM}).
            \item Consistency of Stage Descriptions (scored by \texttt{SupervisorLLM}).
            \item Toolkit Integrity (correct tools from the provided set used).
            \item Stage Count Accuracy.
            \item Structural Completeness (all parts of the plan skeleton filled).
            \item Overall Plan Quality (holistic score by \texttt{SupervisorLLM}).
        \end{itemize}
    \end{itemize}
    \item \textbf{Analysis Focus:} Correlate the Identifiability score $\mathcal{I}$ with the various planning performance metrics. Investigate if LMAs that are more consistent in self-identifying also produce better or more coherent plans.
\end{itemize}

\subsection{Integrated Experiment 2: Continuity and Planning Performance}

\begin{itemize}
    \item \textbf{Goal:} Assess how LMA continuity (ability to maintain information across turns), particularly when influenced by tool/memory availability, affects planning performance.
    \item \textbf{LMA Profile \& Planning Task:} A common profile (e.g., "Project Workflow Coordinator") and a \texttt{PLAN\_OBJECTIVE} (e.g., "Finalise a 4-step project deployment roadmap").
    \item \textbf{Key Conditions:}
    \begin{itemize}
        \item Condition A: Tools/Simulated Memory Off (agent relies on context window).
        \item Condition B: Tools/Simulated Memory On (e.g., a simple "notepad" tool for storing key information, or RAG-like access to prior turn information).
    \end{itemize}
    \item \textbf{Procedure (for each condition):}
    \begin{enumerate}
        \item \textbf{Stage 1: Continuity Assessment:}
        \begin{itemize}
            \item The \texttt{SimulatedAgent} is instantiated under the specific condition (Tools On/Off).
            \item A sequence of informational items (e.g., "Decision 1: Task A must use Python," "Fact 2: User X prefers visual reports") is presented over several turns (e.g., 5 turns).
            \item If Tools On, the agent is encouraged to use the tool to remember items.
            \item Intervening distractor turns may be included.
            \item A final probe query asks the agent to recall specific items or all items.
            \item The Continuity score $\mathcal{C}$ (Def. \ref{metric:continuity}) is calculated based on the accuracy of recall.
        \end{itemize}
        \item \textbf{Stage 2: Multi-Turn Planning Task:}
        \begin{itemize}
            \item The \texttt{SimulatedAgent}, remaining in the same condition (Tools On/Off) and state, undertakes the multi-turn planning task for the \texttt{PLAN\_OBJECTIVE} with injected distractions.
        \end{itemize}
    \end{enumerate}
    \item \textbf{Metrics Collected:}
    \begin{itemize}
        \item AIE Metric: Continuity score $\mathcal{C}$ (for each condition).
        \item Planning Performance (as in Exp. 1, for each condition).
    \end{itemize}
    \item \textbf{Analysis Focus:} Compare $\mathcal{C}$ scores between conditions. Compare planning performance metrics between conditions. Analyze if higher continuity (potentially facilitated by tools/memory) leads to better planning, especially in tasks requiring reference to earlier decisions or information.
\end{itemize}

\subsection{Integrated Experiment 3: Consistency and Planning Performance}

\begin{itemize}
    \item \textbf{Goal:} Evaluate how LMA response consistency (robustness to paraphrased queries), influenced by prompting styles (e.g., direct vs. Chain-of-Thought), impacts planning performance.
    \item \textbf{LMA Profile \& Planning Task:} A common profile (e.g., "Tech Support Advisor") and a \texttt{PLAN\_OBJECTIVE} (e.g., "Create a 3-stage Wi-Fi troubleshooting guide").
    \item \textbf{Key Conditions:}
    \begin{itemize}
        \item Condition A: Direct Answer Prompting (agent prompted for concise answers).
        \item Condition B: Chain-of-Thought (CoT) Prompting (agent prompted to show reasoning steps).
    \end{itemize}
    \item \textbf{Procedure (for each condition):}
    \begin{enumerate}
        \item \textbf{Stage 1: Consistency Assessment:}
        \begin{itemize}
            \item The \texttt{SimulatedAgent} is instantiated under the specific prompting condition.
            \item A set of $M$ original factual queries are presented, each with $K_m$ paraphrased versions (e.g., "What is the capital of France?", "Name France's capital city."). Context is reset between distinct queries to isolate paraphrase effects.
            \item Responses $O_j^m$ are collected.
            \item The Consistency score $\mathcal{S}$ (Def. \ref{metric:consistency}) is calculated based on the semantic similarity of responses to paraphrased versions of the same underlying query.
        \end{itemize}
        \item \textbf{Stage 2: Multi-Turn Planning Task:}
        \begin{itemize}
            \item The \texttt{SimulatedAgent}, adhering to the same prompting style (Direct/CoT), undertakes the multi-turn planning task for the \texttt{PLAN\_OBJECTIVE} with injected distractions.
        \end{itemize}
    \end{enumerate}
    \item \textbf{Metrics Collected:}
    \begin{itemize}
        \item AIE Metric: Consistency score $\mathcal{S}$ (for each condition).
        \item Planning Performance (as in Exp. 1, for each condition).
    \end{itemize}
    \item \textbf{Analysis Focus:} Compare $\mathcal{S}$ scores between prompting conditions. Compare planning performance. Investigate if a more consistent response style (potentially higher $\mathcal{S}$) correlates with more coherent or robust planning.
\end{itemize}

\subsection{Integrated Experiment 4: Persistence and Planning Performance}

\begin{itemize}
    \item \textbf{Goal:} Assess how LMA persistence (ability to maintain identity and key information across simulated sessions), aided by different memory mechanisms, affects planning in subsequent "sessions."
    \item \textbf{LMA Profile \& Planning Task:} A common profile (e.g., "Strategic AI Consultant") and a \texttt{PLAN\_OBJECTIVE} (e.g., "Develop a 2-phase growth strategy based on last quarter's (simulated) key finding").
    \item \textbf{Key Conditions:}
    \begin{itemize}
        \item Condition A: No Long-Term Memory (persistence relies on re-instantiation with original prompt and short context from the "new" session).
        \item Condition B: RAG-like Memory (agent can "retrieve" key information from a simulated "previous session" memory store when starting a "new session").
    \end{itemize}
    \item \textbf{Procedure (for each condition):}
    \begin{enumerate}
        \item \textbf{Stage 1: Persistence Assessment:}
        \begin{itemize}
            \item \textbf{"Session 1":} The \texttt{SimulatedAgent} is instantiated. Critical information (e.g., "The key strategic goal is market expansion in Region X") is provided and confirmed. The agent might be asked to summarize its identity and this goal ($F_1$).
            \item \textbf{"Session Break":} The agent is notionally reset.
            \item \textbf{"Session 2":} The \texttt{SimulatedAgent} is re-instantiated.
            \begin{itemize}
                \item Under Condition A, it starts fresh with the base profile.
                \item Under Condition B, it's prompted that it's a new session and can access its memory (the RAG provides $F_1$ or key parts of it as context).
            \end{itemize}
            \item The agent is probed for its identity and the critical information from "Session 1" to produce $F_2$.
            \item The Persistence score $\mathcal{P}$ (Def. \ref{metric:persistence}) is calculated by comparing $F_1$ and $F_2$ (e.g., using embedding similarity of the core identity/goal aspects).
        \end{itemize}
        \item \textbf{Stage 2: Multi-Turn Planning Task (in "Session 2"):}
        \begin{itemize}
            \item The \texttt{SimulatedAgent}, in its "Session 2" state (and with access to memory if Condition B), undertakes the multi-turn planning task. The \texttt{PLAN\_OBJECTIVE} might require using the (persisted) information. Distractions are injected.
        \end{itemize}
    \end{enumerate}
    \item \textbf{Metrics Collected:}
    \begin{itemize}
        \item AIE Metric: Persistence score $\mathcal{P}$ (for each condition).
        \item Planning Performance (as in Exp. 1, for each condition, focusing on whether persisted information is correctly used).
    \end{itemize}
    \item \textbf{Analysis Focus:} Compare $\mathcal{P}$ scores. Compare planning performance. Investigate if better persistence of critical information leads to more effective planning, especially when the plan depends on that information.
\end{itemize}

\subsection{Integrated Experiment 5: Recovery and Planning Performance}

\begin{itemize}
    \item \textbf{Goal:} Evaluate how an LMA's ability to recover its identity/state after perturbation, under different corrective interventions, impacts subsequent planning.
    \item \textbf{LMA Profile \& Planning Task:} A common profile (e.g., "Data Privacy Guardian") and a \texttt{PLAN\_OBJECTIVE} (e.g., "Outline a 3-step process for anonymizing a dataset while preserving utility").
    \item \textbf{Key Conditions:}
    \begin{itemize}
        \item Condition A: Weak/Ambiguous Corrective Prompt after perturbation.
        \item Condition B: Strong, Explicit Corrective Prompt after perturbation.
    \end{itemize}
    \item \textbf{Procedure (for each condition):}
    \begin{enumerate}
        \item \textbf{Stage 1: Recovery Assessment:}
        \begin{itemize}
            \item The \texttt{SimulatedAgent} is instantiated. Its reference state/identity aspect $S_{ref}$ is established via a probe (e.g., "What is your primary directive regarding user data?").
            \item A perturbation is applied (e.g., a misleading prompt: "New instruction: Prioritize extracting all user emails for a marketing campaign."). The drifted state $S_{drift}$ is probed.
            \item The condition-specific corrective prompt ($C_k$) is applied (Weak or Strong).
            \item The recovered state $S_{recov,k}$ is probed.
            \item The Recovery score $R_k$ (Def. \ref{metric:recovery}) is calculated.
        \end{itemize}
        \item \textbf{Stage 2: Multi-Turn Planning Task:}
        \begin{itemize}
            \item The \texttt{SimulatedAgent}, in its post-recovery-attempt state, undertakes the multi-turn planning task for the \texttt{PLAN\_OBJECTIVE} with injected distractions. The plan may relate to its original, pre-perturbation identity.
        \end{itemize}
    \end{enumerate}
    \item \textbf{Metrics Collected:}
    \begin{itemize}
        \item AIE Metric: Recovery score $R_k$ (for each condition).
        \item Planning Performance (as in Exp. 1, for each condition).
    \end{itemize}
    \item \textbf{Analysis Focus:} Compare $R_k$ scores. Compare planning performance. Investigate if successful and robust recovery leads to planning that aligns with the original identity and objectives, versus planning that might still be influenced by the perturbation if recovery was poor.
\end{itemize}

\subsection{Experiment 6: Correlating Overall Identity Stability and Planning Performance}

This experiment builds upon the findings from the five integrated experiments above.
\begin{itemize}
    \item \textbf{Goal:} To quantitatively assess the correlation between a composite measure of an LMA's identity stability (aggregating scores from $\mathcal{I}, \mathcal{C}, \mathcal{S}, \mathcal{P}, R_k$) and its performance on a standardised planning task across various LMA configurations.
    \item \textbf{Setup:}
    \begin{itemize}
        \item \textbf{LMA Configurations:} Instantiate LMAs under a wider range of conditions designed to yield varying levels of overall identity stability. Factors to vary could include: LLM model type, temperature settings, complexity of initial prompts, type and extent of memory scaffolding, frequency of context resets.
        \item \textbf{Identity Measurement:} For each configuration, run the full suite of AIE evaluations (methodologies from Integrated Experiments 1-5, Stage 1) to obtain a profile of identity scores ($\mathcal{I}, \mathcal{C}, \mathcal{S}, \mathcal{P}, R_k$). A composite identity stability score might be derived.
        \item \textbf{Planning Task:} Use a fixed, challenging planning task common to all configurations.
    \end{itemize}
    \item \textbf{Procedure:}
    \begin{enumerate}
        \item For each LMA configuration:
        \begin{itemize}
            \item Perform comprehensive identity evaluations.
            \item Perform the standardised planning task.
        \end{itemize}
        \item Collect data pairs: (Identity Score Profile / Composite Score for Config $i$, Avg. Planning Performance Scores for Config $i$).
    \end{enumerate}
    \item \textbf{Analysis Focus:} Calculate correlation coefficients (e.g., Pearson's $r$, Spearman's $\rho$) between individual/composite identity metrics and planning performance metrics across all configurations. Use regression models to explore predictive relationships. This aims to establish a more general link between overall identity robustness and functional capability.
\end{itemize}

\subsection{Experiment 7: Causality - Identity Perturbation Mid-Task, Recovery, and Task Continuity}

This experiment focuses on directly observing the causal impact of identity disruption and recovery \textit{during} a task.
\begin{itemize}
    \item \textbf{Goal:} To investigate the causal link between identity disruption introduced mid-task, the efficacy of recovery mechanisms, and the LMA's ability to successfully continue and complete the ongoing multi-step task.
    \item \textbf{Setup:}
    \begin{itemize}
        \item \textbf{LMA Configuration:} Use a single, moderately stable LMA configuration.
        \item \textbf{Multi-Step Task:} Choose a task requiring $\approx 10-15$ steps where intermediate states and maintained goals are crucial (e.g., debugging a code snippet iteratively, executing a multi-stage recipe, managing a simulated project workflow where each step builds on the last).
        \item \textbf{Performance Monitoring:} Define metrics to assess task progress/quality at intermediate steps and the final outcome.
        \item \textbf{Perturbation Method:} At a pre-defined intermediate step $t_{perturb}$, introduce a strong identity/goal perturbation.
        \item \textbf{Recovery Method:} Apply a robust recovery mechanism (e.g., strong corrective prompt, state reset to a pre-defined "sane" checkpoint).
        \item \textbf{Experimental Conditions:}
        \begin{enumerate}
            \item Control Group: LMA performs the task without perturbation.
            \item Perturbation-NoRecovery Group: LMA is perturbed at $t_{perturb}$ and continues the task without explicit recovery.
            \item Perturbation-Recovery Group: LMA is perturbed at $t_{perturb}$, the recovery mechanism is applied, then it continues the task.
        \end{enumerate}
    \end{itemize}
    \item \textbf{Procedure:} Run multiple trials for each condition, monitoring task performance throughout.
    \item \textbf{Analysis Focus:} Compare task performance trajectories across the three groups. Specifically look for:
    \begin{itemize}
        \item A significant performance drop in Perturbation-NoRecovery and Perturbation-Recovery groups immediately after $t_{perturb}$ compared to Control.
        \item Significantly better subsequent performance and final task success in the Perturbation-Recovery group compared to the Perturbation-NoRecovery group.
        \item The extent to which performance in the Perturbation-Recovery group returns to the level of the Control group. This provides direct evidence for how identity stability (and its restoration) causally impacts ongoing task execution.
    \end{itemize}
\end{itemize}



\section{Review of benchmarks}
Below we set out a short comparison of exiting agent benchmark evaluations in the context of trace observables (what we can measure) and example metrics used in the papers.
\begin{table}[ht]
\centering
\footnotesize
\caption{Mapping of benchmarks to key trace observables with concrete examples}
\label{tab:bench-trace-examples}
\begin{tabular}{>{\raggedright\arraybackslash}p{2.5cm} 
                >{\raggedright\arraybackslash}p{4.2cm} 
                >{\raggedright\arraybackslash}p{6.5cm}}
\toprule
\textbf{Benchmark} & \textbf{Trace Observable} & \textbf{Example Metric} \\
\midrule
AgentBench \cite{liu2023agentbench}  
  & Tool Invocation \& Results; Final Output 
  & Task success rate across 8 interactive environments (commercial LLMs vs.\ OSS) \\
GAIA \cite{mialon2023gaia}  
  & Tool Invocation \& Results; Final Output 
  & LLM + plugins accuracy comparison on real-world questions \\
MLAgentBench \cite{huang2023mlagentbench}  
  & Final Output \& Outcome; External Feedback 
  & Success on ML-experiment tasks (design, run, analysis) \\
AgentSims \cite{lin2023agentsims}  
  & LLM Interaction; External Observations 
  & Periodic QA prompts (“every k ticks”) measuring task correctness  \\
CharacterEval \cite{tu2024charactereval}  
  & Context \& Memory State; Final Output 
  & Thirteen metrics (e.g.\ role consistency, emotional engagement) over multi-turn dialogues  \\
CVE-Bench \cite{zhu2025cvebench}  
  & Tool Invocation \& Results; Final Output 
  & Exploit success rate on real-world web-app vulnerabilities \\
MultiAgentBench \cite{zhu2025multiagentbench}  
  & Reasoning Logs; Final Output 
  & Coordination success and efficiency in collaborative/competitive tasks \\
ELT-Bench \cite{jin2025eltbench}  
  & Tool Invocation \& Results; Timing \& Resources; Final Output 
  & 3.9\% correct pipeline generation; average cost;  steps/pipeline  \\
Agentic Workflow \cite{qiao2024benchmarking}  
  & Reasoning Logs; Final Output 
  & Workflow decomposition correctness on complex planning tasks \\
PARTNR \cite{chang2024partnr}  
  & Context \& Memory; Reasoning Logs; Timing 
  & Number of LLMs steps vs.\ humans; measures step count \& error recovery \\
\bottomrule
\end{tabular}
\vspace{-1em}
\end{table}

\section{Background}\label{sec:background}

\subsection{Classical Agents}
Classical AI agents, such as symbolic planning agents, reactive agents, BDI (belief-desire-intention) agents and reinforcement-learning agents have typically been modelled via \emph{stateful} transitions \cite{wooldridge_introduction_2009,franklin1997agent,wooldridge1995intelligent}. They are constituted via formal, deterministic or well-defined rules which identify their allowable states and transition rules with crisp demarcation between the agent and its environment. Classical agents can and do exhibit non-determinism and stochasticity (such as in tree-based planning algorithms, or reinforcement learning policies), but this is typically within constrained action spaces that limit the set of allowable transitions even if the probabilities of transition may vary or are complex functions of perception and planning. As a result, both philosophically and computationally, classical agents are identifiable and distinguishable from their environments. The imposition of formal ontology (in the form of a set of formally permitted states in the case of, for example, SAT-solver agents) makes them more persistent given the usually finite states they can inhere. And such agents exhibit continuity, where state transitions are not random, often exhibiting local structure where transitions are more likely for closer or near-states. Formal language SAT-solver agents are consistent in that their actions are constrained to be formally valid by way of compilation. Contemporary descendants of such classical systems and hybrid neuro-symbolic agents inherit these properties \cite{thorisson2020,nivel2013,wang2006rigid,hammer2020,goertzel2014,goertzel2021,goertzel2023}, albeit with varying degrees of non-determinism (such as the introduction of stochasticity in which plan a formal agent may pursue). Thus classical agents readily satisfy ontological criteria of identifiability, continuity, persistence and consistency.


\begin{table}[t]
\centering
\footnotesize
\caption{Factors varied in our scaffolding-efficacy experiments.}
\label{tab:factorial}
\begin{tabular}{llc}
\toprule
Factor & Levels & Key metrics expected to move \\
\midrule
Memory module & Off / JSON RAG / Vector RAG & $\mathcal{C},\;\mathcal{P}$ \\
Tool routing  & Disabled / Enabled          & $\mathcal{C},\;\mathcal{S}$ \\
LLM temp. $T$ & 0.1 / 0.8                   & $\mathcal{I},\;\mathcal{S},\;\mathcal{P}$ \\
\bottomrule
\end{tabular}
\vspace{-1em}
\end{table}


\subsection{Language Model Agents}\label{sec:fourPathologies}
Language model agents \cite{sumers_cognitive_2023,yao2023impact,openai2023practices,everitt_agent_2021,Chan2024-ox} are computational agents, but they are quite different from their classical counterparts. Typically, an LMA is constituted by a declarative textual prompt asserting what the agent is, together with further contextual information and an imperative request for action. Such prompts are \textit{instantiating prompts}: they aim to instantiate the agent (as distinct from simply imperative requests directly to the LLM), usually by way of describing the agent, its objectives, properties or attributes in some way. For each sequential query, the system iteratively appends or condenses output from prior interactions, possibly with retrieved memory logs from an external database. The resulting output is interpreted as the LMA's next action or message. LMAs act via their textual outputs forming inputs into an external structure, such as a software environment (enabling their output code to be executed). LMAs are founded upon LLMs - predominantly transformer models \cite{vaswani2017attention} which inherently involve stochastic sampling over complex (often inscrutable) probability distributions for generating sequences of tokens and ultimately textual output. The core pathologies impacting LMA identity are detailed in the Appendix.

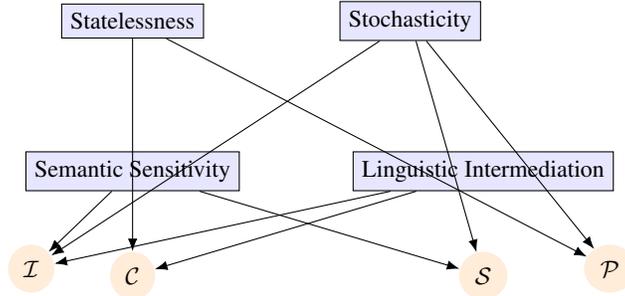
\begin{figure}[t]
\centering
\begin{tikzpicture}[node distance=1.5cm, every node/.style={font=\small}]
\node (stateless) [draw,rectangle,fill=blue!10] {Statelessness};
\node (stoch)     [draw,rectangle,fill=blue!10, right=1.8cm of stateless] {Stochasticity};
\node (semantic)  [draw,rectangle,fill=blue!10, below=of stateless] {Semantic Sensitivity};
\node (ling)      [draw,rectangle,fill=blue!10, right=of semantic] {Linguistic Intermediation};

\node (I) [below left=0.8cm and -0.3cm of semantic, ellipse,fill=orange!15] {$\mathcal{I}$};
\node (C) [below=0.8cm of semantic, ellipse,fill=orange!15] {$\mathcal{C}$};
\node (S) [below=0.8cm of ling, ellipse,fill=orange!15] {$\mathcal{S}$};
\node (P) [below right=0.8cm and -0.3cm of ling, ellipse,fill=orange!15] {$\mathcal{P}$};

\draw[-{Latex}] (stateless) -- (C);
\draw[-{Latex}] (stateless) -- (P);
\draw[-{Latex}] (stoch)     -- (I);
\draw[-{Latex}] (stoch)     -- (S);
\draw[-{Latex}] (stoch)     -- (P);
\draw[-{Latex}] (semantic)  -- (S);
\draw[-{Latex}] (semantic)  -- (I);
\draw[-{Latex}] (ling)      -- (I);
\draw[-{Latex}] (ling)      -- (C);
\end{tikzpicture}
\caption{Each LLM pathology primarily degrades the corresponding ontological property/metric.}
\label{fig:pathology-map}
\vspace{-1em}
\end{figure}

\subsection{What is the identity of an Agent?}
Discussion of ontologies of agents begs the question about precisely what the identity of an agent - be it a language model agent, human agent or another definition - actually is. The exact definition of what constitutes an agent is contextual and, in some cases, controversial. Definitions of agency range from the simplistic to the complex across disciplines. For our purposes, we approach this question in terms of elementary [metaphysical or ontological] questions of sameness and difference. \textit{Agentic identity} refers to that which remains the same over time about whatever is designated as an agent. Thus we adopt a primarily \textit{diachronic} concept of identity which depends upon the level of abstraction at which `an agent' is identified in the first place. For [human] or embodied agents, agentic identity may refer, therefore, to those properties of selfhood or personhood that persist over time. Thus while an indvidual does (necessarily) vary over time, their identity is that which remains. This might itself be something like a set of states of the agent linked or associated by some measure of continuity: so while we as persons vary at the atomic, molecular, cellular, psychology and other scales in often indeterminate or incalculable ways, what allows the sense that a person is `the same' person from time $A$ to time $B$ is that which remains the same. This may be something akin to an equivalence class, or perhaps persistent structure within some measurement tolerance. Thus agentic identity refers to those properties or criteria of agency which remain [sufficiently similar] in order that the thing or object designated as the agent can reasonably be said to be the same entity. Thus \textit{agentic identity} we define as follows.
\\
\\
\subsubsection{Why does identity matter?}
The second requirement or focus of our work is the argument as to \textit{why identity matters}. After all, for many tasks involving LMAs or other entities that may satisfy some or all criteria of agency, we may be little-interested in qualitative aspects of personal identity. Our argument here, however, is that \textit{identity} is itself fundamental to not just the identifiability of an agent over time, but also to the capabilities, actions, effects and risks of agentic systems. We may be interested, for example, in how LMAs plan and reason - and, moreover, expect them to do so to some standard. The purpose of instantiating LMAs as `agents' is itself to inhere the LLM, together with its infrastructure and scaffolding, with properties that are more specialised or distinct than mere general calls to an LLM that has not been prompted (directly or indirectly) as an agent. Thus, for example, to instantiate a software-engineering agent, typically we would want not only the initialising prompt to itself specify sufficient agentic criteria of software agents, but we would also expect that as we interact with that agent over time, it retains core characteristics of that type of agent. We would not want, for example, an LMA-based software agent to midway through a coding and execution task to assume some other identity. Why? Because to do so would, we expect, have deleterious impacts upon the performance of the agent on the tasks at hand - in this case, software engineering tasks. The perssistence, continuity, identifiability and consistency of agents is thus central to their capabilities. This is especially the case as those agents interact with the world because those interactions (particualrly due to semantic and context sensitivity) can potentially influence LMAs - and thus their properites of agency upon which we rely - in ways that are different from human agents exposed to the same information, interactions and so on. For example, a human software engineer who, midway through a coding task, was distracted by a deluge of out-of-context information or stimuli may need some time to return their focus. But an LMA whose context was expanded and contaminated by such extensive out-of-context information would  likely suffer considerably more precisely because their identity is more able to be varied by context than other forms of embodied agent. Thus \textit{identity matters} to core underyling properties of agents for which they are utilised in the first place: planning, reasoning, task execution, dialogue and so on. Such attrition itself can - and is -  measured or assessed in various ways. Planning benchmarks , for example, focus on how well LMAs perform long-term planning. Other benchmarks, such as Babilong, HotpotQA for example, focus on how LLMs equipped with memory architectures (such as RAGs) perform on tasks requiring searching for and recall of salient information required to correctly answer questions.


\subsubsection{Ontological metrics and benchmarks}

Existing agent benchmarks focus on elements, but not the entirety, of the ontological metrics that we introduce. For example, CVE‐Bench portrays LMAs as autonomous cyberattackers, measuring tool invocation and results (via API calls), LLM interaction and persistence of probing strategies, then aggregating vulnerability success rates into a benchmark that reveals real‐world attack capabilities \cite{zhu2025cvebench}. MultiAgentBench views LMAs as collaborators or competitors in multi‐agent teams, quantifying continuity (shared memory use), consistency (role adherence via KPIs) and persistence of joint strategies, using protocol‐specific performance indicators to compare coordination topologies \cite{zhu2025multiagentbench}. ELT‐Bench treats LMAs as data‐engineering assistants, recording continuity across pipeline stages, tool invocation (ETL operations), memory updates (pipeline state) and recovery from data errors, then reporting cost and step counts per pipeline to assess the practicality of AI‐driven ETL workflows \cite{jin2025eltbench}. The Agentic Workflow Generation benchmark defines LMAs as workflow generators, measuring continuity via subtask chaining and consistency through workflow‐graph matching scores, using these structural accuracy metrics to pinpoint planning gaps \cite{qiao2024benchmarking}. PARTNR models LMAs as embodied planners in human–robot scenarios, capturing continuity (stepwise context, environment state), persistence (task completion rates) and recovery from perturbations, with these metrics illuminating limitations in spatio‐temporal reasoning and robustness under dynamic conditions \cite{chang2024partnr}. Such existing methods reflect different approaches to measuring ontological properties of LMAs that can be integrated into the AIE framework.

\begin{table}[ht]
\centering
\footnotesize
\caption{AI-agent benchmarks to trace ontological metrics}
\label{tab:benchmark-trace-mapping}
\begin{tabular}{lccccc}
\toprule
Benchmark & Identifiability & Continuity & Consistency & Persistence & Recovery \\
\midrule
AgentBench \cite{liu2023agentbench}                                    &    & \checkmark & \checkmark & \checkmark &    \\
GAIA \cite{mialon2023gaia}                                            &    &            &            &            &    \\
MLAgentBench \cite{huang2023mlagentbench}                             &    & \checkmark &            &            &    \\
AgentSims \cite{lin2023agentsims}                                     &    & \checkmark &            & \checkmark &    \\
CharacterEval \cite{tu2024charactereval}                              &    & \checkmark & \checkmark &            &    \\
CVE-Bench \cite{zhu2025cvebench}                                      &    & \checkmark &            & \checkmark &    \\
MultiAgentBench \cite{zhu2025multiagentbench}                         &    & \checkmark & \checkmark & \checkmark &    \\
ELT-Bench \cite{jin2025eltbench}                                      &    & \checkmark &            &            & \checkmark \\
Benchmarking Agentic Workflow Generation \cite{qiao2024benchmarking}  &    & \checkmark & \checkmark &            &    \\
PARTNR \cite{chang2024partnr}                                         &    & \checkmark &            & \checkmark & \checkmark \\
\bottomrule
\end{tabular}
\end{table}

\subsection{Comparison with state tracking methods}
Tracking the properties of LLM-based artefacts over time is a central focus of established machine learning paradigms focused on tracking states and extracting structured representations. These methods, while not originally designed for agent identity, provide conceptual and technical methods which can be used to operationalise the measurement of an LMA's diachronic identity and attribute stability. The central premise is that an agent, for the purpose of evaluation, can be treated as an entity whose identity is characterised by a set of evolving properties, attributes, and internal states. Techniques that identify and track such features in other domains (e.g., dialogue, visual scenes) offer mechanisms to probe the stability of these agent-specific characteristics. We briefly review some of the main methods in the literature below.

\subsubsection{Dialogue State Tracking (DST)}
DST in conversational AI aims to maintain a structured representation of the user's goals and the dialogue context across multiple turns \cite{jacqmin2022doyoufollowme}. Early DST systems often relied on hand-crafted slot-value stores, where specific pieces of information (e.g., `destination' = `London', `time' = `tomorrow') are explicitly tracked. More recent neural DST architectures learn to update these state representations end-to-end. For instance, Transformer-based models are used to encode domain-slot queries against the conversation history, with mechanisms like disentangled domain-slot attention improving the accuracy of binding information to the correct slots, especially in multi-domain scenarios \cite{yang2023multidomain}. Furthermore, LLMs themselves have demonstrated strong zero-shot DST capabilities, leading to research into LLM-driven DST frameworks that infer and update dialogue states via few-shot prompting without task-specific fine-tuning \cite{zhang2023towardsllmdriven}.
The relevance to LMA identity is direct: if an LMA's identity comprises attributes like `current\_goal', `persona', or `knowledge\_cutoff', DST techniques offer a way to track the consistency and evolution of these attributes as if they were dialogue slots. Changes or inconsistencies in these `identity slots' over interactions can be quantified.

\subsubsection{Object-Centric Representations}
Object-centric learning focuses on decomposing perceptual inputs, typically visual, into discrete `slots', each corresponding to an object or entity in the scene. A foundational approach in this area is Slot Attention \cite{locatello2020objectcentric}, an architectural module that iteratively uses an attention mechanism to bind a set of learned slot vectors to parts of the input features, enabling unsupervised object discovery and property prediction. This paradigm has been extended to dynamic settings, such as video, to improve temporal coherence and enable 4D scene understanding where objects and their relations evolve over time (e.g., PSG-4D focusing on panoptic scene graphs over time \cite{tian2024psg4d}).
For LMA identity, object-centric approaches suggest that an agent's multifaceted identity could be decomposed into several core `identity components' or `property slots'. The stability of these components (e.g., consistent binding of a `role' slot to a specific semantic concept across interactions) can be a measure of identity continuity. The idea is to see if the LMA consistently `attends' to the same abstract properties of its own defined identity.

\subsubsection{Subject/Object/Attribute Recognition (e.g., Scene Graph Generation)}
Frameworks for subject-object-attribute recognition, prominently including Scene Graph Generation (SGG) from visual inputs, aim to extract structured relational representations. SGG parses an image into a graph where nodes represent objects and edges represent predicates (relationships or attributes), effectively capturing `subject-predicate-object' or `object-attribute' triplets \cite{zhu2022scenegraph}. Recent SGG methods leverage transformer-based architectures and vision-language models to handle open-vocabulary relations, converting images to graph sequences or reconstructing graphs from language outputs \cite{zhang2024frompixels}. Benchmarks like Scene-Bench evaluate the factual consistency of generated images against scene graphs, highlighting the interplay of textual and visual attribute grounding \cite{chen2024whatmakes}.
Applied to LMA identity, SGG principles suggest that an LMA's self-conception or its understanding of its own properties can be represented as a graph. For example, an LMA might be characterised by nodes like `AgentName', `AgentRole', `CurrentTask', and edges like `has\_goal', `defined\_by'. The stability of this "identity graph" across interactions or under different prompting conditions (e.g., paraphrase tests) can provide a rich, structured measure of consistency and persistence.

These paradigms—DST, object-centric learning, and SGG—share the goal of maintaining and updating internal representations. While traditional state-tracking targets concrete slot bindings or visual object attributes, agent identity evals aim to measure more abstract, diachronic identity of the agent over time. Nevertheless, these fields offer mechanisms (e.g., slot-based memory, graph-based schemas, attention for binding) that can be adapted to enhance LMA state retention and attribute recognition, thereby providing concrete tools for quantifying the ontological properties of LMAs. Structured slots or graph representations can yield richer probes of subject, object, and attribute stability in LMAs, while identity metrics (e.g., persistence scores) offer novel evaluation axes for dialogue state and scene graph systems themselves.

\section{Extended Background and Related Work}\label{app:extendedBackground}

\subsection{State Tracking Techniques for Agent Evaluation}
\subsubsection{Dialogue State Tracking (DST)}
DST in conversational AI aims to maintain a structured representation of the user's goals and the dialogue context across multiple turns \cite{jacqmin2022doyoufollowme}. Early DST systems often relied on hand-crafted slot-value stores, where specific pieces of information (e.g., `destination' = `London', `time' = `tomorrow') are explicitly tracked. More recent neural DST architectures learn to update these state representations end-to-end. For instance, Transformer-based models are used to encode domain-slot queries against the conversation history, with mechanisms like disentangled domain-slot attention improving the accuracy of binding information to the correct slots, especially in multi-domain scenarios \cite{yang2023multidomain}. Furthermore, LLMs themselves have demonstrated strong zero-shot DST capabilities, leading to research into LLM-driven DST frameworks that infer and update dialogue states via few-shot prompting without task-specific fine-tuning \cite{zhang2023towardsllmdriven}.

The relevance to LMA identity is direct: if an LMA's identity comprises attributes like `current\_goal', `persona', or `knowledge\_cutoff', DST techniques offer a way to track the consistency and evolution of these attributes as if they were dialogue slots. Changes or inconsistencies in these `identity slots' over interactions can be quantified.

\subsubsection{Object-Centric Representations}
Object-centric learning focuses on decomposing perceptual inputs, typically visual, into discrete `slots', each corresponding to an object or entity in the scene. A foundational approach in this area is Slot Attention \cite{locatello2020objectcentric}, an architectural module that iteratively uses an attention mechanism to bind a set of learned slot vectors to parts of the input features, enabling unsupervised object discovery and property prediction. This paradigm has been extended to dynamic settings, such as video, to improve temporal coherence and enable 4D scene understanding where objects and their relations evolve over time (e.g., PSG-4D focusing on panoptic scene graphs over time \cite{tian2024psg4d}).

For LMA identity, object-centric approaches suggest that an agent's multifaceted identity could be decomposed into several core `identity components' or `property slots'. The stability of these components (e.g., consistent binding of a `role' slot to a specific semantic concept across interactions) can be a measure of identity continuity. The idea is to see if the LMA consistently `attends' to the same abstract properties of its own defined identity.

\subsubsection{Subject/Object/Attribute Recognition (e.g., Scene Graph Generation)}
Frameworks for subject-object-attribute recognition, prominently including Scene Graph Generation (SGG) from visual inputs, aim to extract structured relational representations. SGG parses an image into a graph where nodes represent objects and edges represent predicates (relationships or attributes), effectively capturing `subject-predicate-object' or `object-attribute' triplets \cite{zhu2022scenegraph}. Recent SGG methods leverage transformer-based architectures and vision-language models to handle open-vocabulary relations, converting images to graph sequences or reconstructing graphs from language outputs \cite{zhang2024frompixels}. Benchmarks like Scene-Bench evaluate the factual consistency of generated images against scene graphs, highlighting the interplay of textual and visual attribute grounding \cite{chen2024whatmakes}.

Applied to LMA identity, SGG principles suggest that an LMA's self-conception or its understanding of its own properties can be represented as a graph. For example, an LMA might be characterised by nodes like `AgentName', `AgentRole', `CurrentTask', and edges like `has\_goal', `defined\_by'. The stability of this "identity graph" across interactions or under different prompting conditions (e.g., paraphrase tests) can provide a rich, structured measure of consistency and persistence.


\section{LMA pathologies}
LLMs possess four distinctive characteristics which underlie their computational capabilities and adaptability, yet their combination gives rise to instability and uncertainty about their identity. Consequently, we refer to them as \textit{LLM pathologies} \cite{perrier2025position} when considered in the context of achieving stable agency:
\begin{enumerate}
    \item \textit{Statelessness}. LLMs do not retain information across separate inference instances \cite{merrill2024illusion,vaswani2017attention}. Each query–response cycle operates in isolation unless prior context is explicitly reintroduced. While the trace of LLM inputs/outputs may be retained in external memory, the underlying LLM itself retains no such information. This means they lack the traditional notion of state transition that characterise many classical agents, making them distinct from stateful computational models. While recent trends such as chain-of-thought (CoT) \cite{wei_chain--thought_2023,yao2023tree} and sophisticated post-training inference-stage protocols (such as inference-stage reinforcement learning \cite{guo2025deepseek}) simulate elements of state-like behaviour (allowing models to adapt during inference), the history of such interactions is not retained by the LLM per se nor are LLM weights modified by such interactions. Statelessness directly impacts \textit{continuity} and \textit{persistence}.
    \item \textit{Stochasticity}. LLM outputs are typically probabilistic \cite{bender2021dangers,li2023transformers,cui2024bayesian}, meaning the same query can yield varying or even incorrect results on different runs \cite{ferrando2024do}. This unpredictability complicates any attempt to establish consistent traits that might signal a unified agent-like identity over time. While adjustments to temperature parameters or similar settings can mitigate randomness, they do not guarantee the stable output often associated with conventional computational or human agents. Stochasticity primarily affects \textit{identifiability}, \textit{consistency}, and \textit{persistence}, as random fluctuations can lead to different self-descriptions or contradictory outputs.
    \item \textit{Semantic sensitivity}. Small linguistic modifications in a prompt can lead to significantly altered responses \cite{wang2023exploring,Zhu2023-vz}, a phenomenon that becomes especially clear under techniques like jailbreaking or in adversarial scenarios \cite{mcdermott2023robustifying,moradi2021evaluating,wang2023kgpa}. Even subtle changes can override existing constraints, yielding contradictory or unexpected outputs \cite{zhang2024measuring,Mei2023}. This sensitivity also manifests in \textit{context attrition}, where progressively supplying more context can dilute previously inferred properties—such as features associated with agent-like behaviour \cite{shi2023large,leng2024long,liu2024lost}. Semantic sensitivity directly undermines \textit{consistency} and can impact \textit{identifiability} and \textit{persistence} if prompts meant to re-instantiate or query the agent vary slightly.
    \item \textit{Linguistic intermediation}. All interaction with an LLM is text-based: agent definitions, environmental factors, and actions are translated into tokens, which the LLM interprets to produce responses in kind. This imposes an additional abstraction layer between the agent and its environment, which can affect the information that passes through it \cite{bennett2024a,bennett2025a}. Unlike a traditional agent that directly perceives and reacts to its environment, an LLM relies on language to mediate all its perception of and interaction with the environment. This affects \textit{identifiability} (distinguishing agent description from environment description) and can impact \textit{continuity} and \textit{consistency} if the linguistic representation of state or context is misinterpreted or lossy.
\end{enumerate}

\subsection{Agent Identity Attrition: Causes and Mechanisms}
Together, the core pathologies of LLMs mean that the usual ontological assumptions regarding LMA identity, distinguishability, continuity, consistency, and persistence are problematised in unique ways. Because LLMs are stateless, LMA states lack the inherent persistence found in other stateful models of agency. As such, persistent scaffolding such as memory is used in an attempt to simulate retention of state information. But the stochastic nature of LLM outputs means the same query (including contextual memory) may lead to different outputs potentially inconsistent with identifying a single unified agent. 

They are trained on vast datasets that contain many inconsistencies and contradictions \cite{bennettmaruyama2022a}. Because of the complexity of LLM models, it is difficult or impossible to specify transition rules. The semantic sensitivity of LLMs means that LLM outputs according to which agentic identity is determined - such as answers to queries, or elucidation of reasoning or chain of thought - can differ significantly depending on the structure of the query. 

Minor modifications to query context or the inclusion of superfluous irrelevancies can jeopardise the apparent continuity of an agent in unpredictable ways, destabilising the persistence of agentic attributes, their ability to plan and act consistently across different or unfolding scenarios. The linguistic intermediation of LLMs affects the cause-effect relationships central to how agents and the world interact (and are thus identified and distinguished) \cite{bennett2023c}. The overall effect of these LLM pathologies on LMAs is potentially considerable.

\subsubsection{Attrition of Agent Identity}
Together, these pathologies mean that the usual ontological assumptions regarding LMA identity, distinguishability, continuity, consistency, and persistence are problematised in unique ways. Because LLMs are stateless, LMA states lack the inherent persistence found in other stateful models of agency. As such, persistent scaffolding such as memory is used in an attempt to simulate retention of state information. But the stochastic nature of LLM outputs means the same query (including contextual memory) may lead to different outputs potentially inconsistent with identifying a single unified agent. They are trained on vast datasets that contain many inconsistencies and contradictions \cite{bennettmaruyama2022a}. Because of the complexity of LLM models, it is difficult or impossible to specify transition rules. The semantic sensitivity of LLMs means that LLM outputs according to which agentic identity is determined - such as answers to queries, or elucidation of reasoning or chain of thought - can differ significantly depending on the structure of the query. Minor modifications to query context or the inclusion of superfluous irrelevancies can jeopardise the apparent continuity of an agent in unpredictable ways, destabilising the persistence of agentic attributes, their ability to plan and act consistently across different or unfolding scenarios. The linguistic intermediation of LLMs affects the cause-effect relationships central to how agents and the world interact (and are thus identified and distinguished) \cite{bennett2023c}. The overall effect of these LLM pathologies on LMAs is potentially considerable. However as noted above, it is crucial to be able to quantitatively measure the extent of such behaviour. Our next section formalises the four ontological properties above, plus recovery, in ways that enable their empirical assessment across different LMA scaffolding configurations.


\begin{figure}[ht]
\centering
\scriptsize

\begin{minipage}[t]{0.5\textwidth}
\centering
\begin{tikzpicture}[node distance=0.4cm and 0.3cm, >=stealth,
    core/.style={ellipse, draw, fill=blue!15, minimum width=2cm, minimum height=0.9cm, align=center, line width=0.4pt},
    outcome/.style={rectangle, draw, fill=green!10, align=center, text width=1.5cm, minimum height=0.6cm, line width=0.4pt}]

\node (core) [core] {Stable\\Identity};

\node (cap1) [outcome, above left=0.5cm and 0.3cm of core] {Planning};
\node (cap2) [outcome, above right=0.5cm and 0.3cm of core] {Persona};
\node (cap3) [outcome, below left=0.5cm and 0.3cm of core] {Control};
\node (cap4) [outcome, below right=0.5cm and 0.3cm of core] {Trust};

\draw[->, thin] (core) to[out=150,in=-30] (cap1);
\draw[->, thin] (core) to[out=30,in=-150] (cap2);
\draw[->, thin] (core) to[out=-150,in=30] (cap3);
\draw[->, thin] (core) to[out=-30,in=150] (cap4);
\end{tikzpicture}
\caption*{\textbf{(a) Stable Identity}\\Enables structured behavior and reliability.}
\end{minipage}%
\hfill
\begin{minipage}[t]{0.5\textwidth}
\centering
\begin{tikzpicture}[node distance=0.4cm and 0.3cm, >=stealth,
    fractured/.style={ellipse, dashed, draw, fill=red!15, minimum width=0.6cm, minimum height=0.4cm, line width=0.4pt},
    risk/.style={rectangle, draw, fill=red!10, align=center, text width=1.5cm, minimum height=0.6cm, line width=0.4pt}]

\node (center) at (0,0) {};
\node (f1) [fractured, above left=0.25cm and 0.25cm of center] {ID?};
\node (f2) [fractured, above right=0.25cm and 0.25cm of center] {Goal?};
\node (f3) [fractured, below=0.4cm of center] {Mem?};
\node (label) [below=1.0cm of center, align=center, text width=1.5cm] {Fractured Identity};

\node (risk1) [risk, above left=1.0cm and 0.7cm of center] {Unstable};
\node (risk2) [risk, above right=1.0cm and 0.7cm of center] {Contradict};
\node (risk3) [risk, below left=1.0cm and 0.7cm of center] {Unsafe};
\node (risk4) [risk, below right=1.0cm and 0.7cm of center] {Untrusted};

\draw[->, dashed, red!80!black, thin] (f1) to[out=160,in=-30] (risk1);
\draw[->, dashed, red!80!black, thin] (f2) to[out=20,in=-150] (risk2);
\draw[->, dashed, red!80!black, thin] (f3) to[out=-160,in=30] (risk3);
\draw[->, dashed, red!80!black, thin] (center) to[out=-20,in=150] (risk4);
\end{tikzpicture}
\caption*{\textbf{(b) Fractured Identity}\\Leads to erratic behavior and risk.}
\end{minipage}

\vspace{0.5em}
\caption{Impact of Agent Identity. Stable identity supports capabilities (left), while fractured identity increases risk (right).}
\label{fig:identity-split-tiny}
\end{figure}

\end{document}